\documentclass[lettersize,journal]{IEEEtran}
\usepackage{amsmath,amsfonts}
\usepackage{algorithmic}
\usepackage{algorithm}
\usepackage{array}
\usepackage{multirow}
\usepackage[caption=false,font=normalsize,labelfont=sf,textfont=sf]{subfig}
\usepackage{textcomp}
\usepackage{stfloats}
\usepackage{url}
\usepackage{verbatim}
\usepackage{graphicx}
\usepackage{cite}
\begin{document}

\title{GeoVLM: Improving Automated Vehicle Geolocalisation Using Vision-Language Matching}

\author{Barkin Dagda$^{1}$, Muhammad Awais$^{2}$, Saber Fallah$^{1}$
\thanks{$^{1}$Barkin Dagda and Saber Fallah are with the Connected and Autonomous Vehicles Lab (CAV-Lab), University of Surrey, Guildford, GU2 7XH, UK.%
        {\tt\small [bd00242, s.fallah@surrey.ac.uk}}
\thanks{$^{2}$Muhammad Awais is with the Centre for Vision, Speech and Signal Processing (CVSSP), University of Surrey, Guildford, GU2 7XH, UK. %
        {\tt\small muhammad.awais@surrey.ac.uk}}%
\thanks{Manuscript received May 5, 2025; revised [REVISE DATE].}
}

\markboth{Journal of \LaTeX\ Class Files,~Vol.~14, No.~8, August~2021}%
{Shell \MakeLowercase{\textit{et al.}}: A Sample Article Using IEEEtran.cls for IEEE Journals}

\IEEEpubid{0000--0000/00\$00.00~\copyright~2021 IEEE}

\maketitle

\begin{abstract}
Cross-view geo-localisation identifies coarse geographical position of an automated vehicle by matching a ground-level image to a geo-tagged satellite image from a database. Despite the advancements in Cross-view geo-localisation, significant challenges still persist such as similar looking scenes which makes it challenging to find the correct match as the top match. Existing approaches reach high recall rates but they still fail to rank the correct image as the top match. To address this challenge, this paper proposes GeoVLM, a novel approach which uses the zero-shot capabilities of vision language models to enable cross-view geo-localisation using interpretable cross-view language descriptions. GeoVLM is a trainable reranking approach which improves the best match accuracy of cross-view geo-localisation. GeoVLM is evaluated on standard benchmark VIGOR and University-1652 and also through real-life driving environments using Cross-View United Kingdom, a new benchmark dataset introduced in this paper. The results of the paper show that GeoVLM improves retrieval performance of cross-view geo-localisation compared to the state-of-the-art methods with the help of explainable natural language descriptions. The code is available at {https://github.com/CAV-Research-Lab/GeoVLM}
\end{abstract}


\section{Introduction}
\IEEEPARstart{L} {ocalisation} in automated vehicles refer to the process of finding the precise position and orientation of the automated system or a robot within a given environment relative to a chosen reference coordinate system \cite{Jia2008AutonomousVN}. Localisation in automated vehicles serves as
a backbone for higher-level functions such as perception,
planning, and control, ensuring the vehicle can navigate safely and effectively. The most common solution for estimating the geo-position of automated vehicles is Global Positioning System (GPS). However, GPS suffers from limitations including signal occlusion, electromagnetic interference and it can be unavailable in certain driving environments (e.g., urban and rural areas with dense forestations). The map aided localisation approaches use high definition representation of the environment and try to match the live sensor measurements (e.g., LiDAR scans) with the prior map. The map based approaches can provide high localisation accuracy but they require expensive storage space and computational power onboard. The cross-view geo-localisation (CVGL) which is a branch of vision based localisation is a substitute method for GPS-denied areas offering a lightweight and adaptable solution by matching a ground level image (also known as query image) captured from the vehicle with a database of geo-tagged aerial images (also known as satellite or reference images). In the domain of automated vehicles, vision based localisation holds visual place recognition (VPR) at its core. VPR is more than a conventional image retrieval task, rather a fundamental challenge of environment recognition that requires the extraction of features invariant to seasonal changes, weather conditions, and illumination variations in previously observed environments. Addressing this challenge starts by formulating the
VPR as an image retrieval task where a collection of geo-
tagged images (database) are needed to serve as a reference.
The nature and precision of these geo-tags vary according
to application requirements. The landmark identifiers are enough for general place recognition tasks but automated vehicle
navigation demands higher precision references, in the form of GPS coordinates. The availability of geo-tagged aerial images through Google Maps and Bing Maps offer a publicly available data source to determine the vehicle's location through cross-view image retrieval task. However, the cross-view image matching is a challenging computer vision task yet to be solved due to: 1) drastic viewpoint differences between cross views 2) seasonal and time frame variations (e.g., day-time, night-time) between aerial and query images 3) dynamic environments (e.g., moving vehicles and pedestrians on the ground level). Previous works approached these challenges from a pure vision perspective to find the correspondences between two viewpoints. Zhu et. al \cite{Zhu2022TransGeo} proposed the first pure Vision Transformer (ViT) based approach which outperformed the Convolutional Neural Network (CNN) based solutions by establishing a long-range correlations and a relationship between cross-view image patches. Recent works such as Zhang et. al \cite{zhang2023crossview} built up on the ViT idea and processed the low-level features along with the global information extracted from a geometric layout. However, these methods increase the complexity of the task by introducing advanced geometric projections between two views. The increased complexity not only makes it difficult to understand the rationale behind the correct matches but also reach limited accuracy for finding correct pair as the top match. Although, the recent approaches achieve high recall rates identifying the correct pair within top-10, they lack the visual reasoning. Their opaque nature also makes it more challenging to understand the knowledge transfer across new geographic landscapes. Furthermore, the real-time applicability of these approaches has not been evaluated in realistic driving conditions, such as crowded highways where feature matching becomes challenging due to sparse features, surrounding vehicles and challenging environmental conditions like low-light environments or seasonal variations.

This paper presents the first cross-view geolocalisation approach \textit{GeoVLM} which improves the state-of-the-art performance by integrating an explainable reranking approach into the cross-view image retrieval task. In this work, vision and language are used in cooperation to pre-generate natural language descriptions for the cross-view images relying on the zero-shot performance of Vision Language Models (VLMs). By using language descriptions along with the image features, GeoVLM combines two modalities to develop visual reasoning for reranking the correct image pair.

The addition of vision and/or sound or other modalities to the large language models (LLMs) create language based large multi-modal models. With advanced developments of LLMs such as ChatGPT \cite{OpenAI2023ChatGPT} and Llama series \cite{Touvron2023Llama}, \cite{Touvron2023Llama2, llama3} the applications of language based large multi modal models become more adaptable and generalisable for automated driving tasks e.g., DriveGPT4 \cite{xu2024drivegpt4} and HighwayLLM \cite{highwayllm} used LLM for motion planning enabling the transparency behind the learning-based actions. Other studies such as Talk2BEV \cite{choudhary2023talk2bev} use VLM to reason the bird's-eye view maps and predict unsafe scenarios. LingoQA and VLAAD \cite{Park_2024_WACV} provide conversational interfaces between humans and automated vehicles. Moreover, frameworks like RAG-Driver \cite{yuan2024ragdriver} use retrieval-augmented generation (RAG) to explain driving actions alongside control signal predictions. In perception tasks, LC-LLM \cite{peng2024lcllm} benefits from the reasoning capabilities of LLMs to enhance lane change prediction with explanations. Similarly, GeoVLM benefits from the human level reasoning capabilities of VLMs to bring explainability and aims to improve the SOTA performance of the cross-view image retrieval task. It benefits from the strong visual reasoning and zero-shot capabilities of foundation models which is the ability to perform new tasks without requiring specific training examples for those tasks.

The recent works such as Sample4Geo \cite{sample4geo} achieve high recall rates in most benchmarks by identifying the correct match in the top-10 for about 90\% of the time which shows a human superior performance. However, the main challenge is to distinguish between the similar looking different geographic regions which require visual understanding, reasoning and comparison of the query image with multiple candidates. For this task, the human way of scene interpretation can be more useful than the learned patterns from the vision models. Therefore, GeoVLM's trainable reranking module uses the semantic descriptions of cross-view scenes which we called as \textit{cross-view language} with their corresponding image embeddings to rerank the top-10 canditates identified by the pretrained cross-view image encoder. Moreover, this paper introduces a new cross-view dataset collected across three major cities of the United Kingdom called the Cross-View UK or the CVUK. The CVUK serves as the first cross-view benchmark dataset which includes different seasons and lighting conditions and includes corresponding scene descriptions for each image (see details in Section \ref{dataset}).
The main contributions of the work are summarized as follows:
\begin{itemize}
    \item To the best of authors' knowledge, the first explainable cross-view geolocalisation approach GeoVLM is introduced. The GeoVLM uses image and cross-view language to train a reranking module for the CVGL task.

    \item To the best of authors' knowledge, the first cross-view dataset is collected in the UK, the CVUK, providing scene descriptions for ground and aerial images across different seasons and lighting conditions, serving as a novel benchmark.

    \item The GeoVLM improves the retrieval accuracy of top-1 and top-5 in VIGOR and the CVUK benchmark.
\end{itemize}
\begin{figure*}[h]
    \centering
    \includegraphics[scale=0.40]{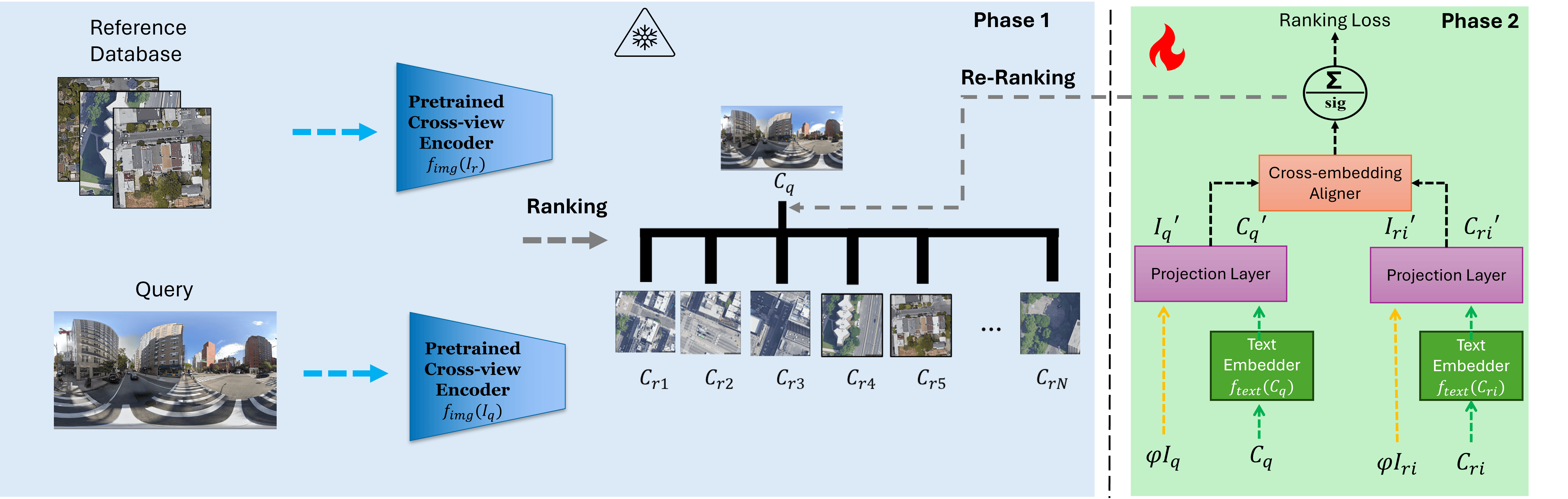}
    \caption{ Overview of proposed GeoVLM pipeline for CVGL. The left side of the figure (blue section) illustrates Phase 1, where the network is frozen and top-k reference images are ranked for each query using pretrained cross-view encoders. The right side (green section) represents Phase 2, the trainable reranking module, where image and text embeddings for each candidate top-k and query image combination are projected into the same latent space and fused to produce a refined ranking score}
    \label{fig:geovlm}
\end{figure*}
\section{RELATED WORK}
CVGL estimates the location of the autonomous vehicle by retrieving the correct geo-tagged aerial view for a given road-level image. Some approaches used the geo-tag as a coarse initial estimate and aimed for more precise positioning accuracy. Shi et al. \cite{shi2022crossview} employed a trainable Siamese CNN pipeline, framing localisation as a neural-network-based pose estimation, while Fervers et al. \cite{fervers2023uncertaintyaware} proposed an uncertainty-aware approach that modeled the vehicle's pose as a probability distribution. Their method utilized a ViT for feature extraction, applying a cross-attention mechanism to transform street-view images into a bird's-eye view, further refining results using self-attention. Wang et al. \cite{hcnet} introduced a correlation-aware homography estimation, iteratively estimating a homography matrix to align the bird's-eye view ground-level image with the satellite image, thereby enhancing fine-grained localisation.

A persistent challenge in this domain is the problem of visually similar scenes \cite{Workman2015WideAreaIG, Ghanem2023LeveragingCG}. In urban environments, infrastructure similarity creates significant localisation challenges. Buildings, streets, and urban landscapes often appear nearly identical from aerial perspectives, complicating precise geo-location identification \cite{Ghanem2023LeveragingCG}. Chen et al. \cite{Chen2025WithoutPL} term this phenomenon "neighbor confusion," where adjacent areas exhibit such high visual similarity that distinguishing between them becomes extremely difficult. This issue is particularly acute in dynamic scenarios involving moving vehicles, where state-of-the-art models struggle to differentiate between consecutive images with minimal visual variation \cite{Ghanem2023LeveragingCG}. Yang et al. \cite{yang2021cross} emphasize that content-based methods must develop sophisticated image representations capable of discriminating between visually similar scenes across diverse environments. Li et al. \cite{Li2024LearningCV} further demonstrate that even advanced foundation models, which perform well in single-view geo-localisation, suffer from significant performance degradation when confronted with cross-view challenges.

Given these difficulties, the majority of CVGL methods focus on improving retrieval-based approaches to enhance generalizability. Shi et al. \cite{safa} introduced polar transformations to bridge the domain gap between two views. Zhu et al. \cite{Zhu2022TransGeo} proposed the first transformer-based approach, leveraging long-range correlations between image patches. Zhang et al. \cite{zhang2023crossview} improved cross-area generalization by separating geometric details from raw features. Their approach learned spatial correlations between cross-views using a geometric layout extractor; however, the authors acknowledged a limitation in fully understanding geometric layout descriptors. Additionally, Tian et al. \cite{TianSMDT} pointed out that existing approaches often neglect environmental changes, such as construction work or seasonal variations, as well as the spatial relationship of co-visibility between views. Their method, SMDT (Semantic segmentation, Mixed polar-perspective mapping, Dual CGAN, and Transformer), incorporated semantic segmentation to model geometric relationships and combined polar and perspective mappings for image alignment. However, the SMDT approach lacks evaluation in cross-area settings.

To further improve generalization and knowledge transferability, Fabian et al. \cite{sample4geo} applied a sampling strategy, proposing a near-neighbor sampling method that selects hard negative images based on their geographical proximity to the query before training. Despite these advancements, CVGL approaches continue to face limitations in achieving high retrieval accuracy and maintaining interpretability. To address these gaps, the proposed GeoVLM integrates image and language features in a trainable re-ranking approach, improving top-match accuracy through visual understanding and reasoning that can be interpreted at a human level. Additionally, this paper quantitatively demonstrates the real-world applicability of CVGL approaches and explores the potential of integrating language-based large multi-modal models in localisation of automated vehicles.

\section{GeoVLM}
This section introduces the proposed GeoVLM approach which uses vision and language for reranking cross-view images.
\subsection{Problem Definition}
Existing CVGL approaches struggle to identify the correct image pair among similar looking scenes which represent different geographical positions. These approaches reach high recall rates ranking the correct satellite image within top-10 but they are limited in the top match accuracy. This requires human level of understanding about the scenes to compare the hidden cues and geometric layouts and reason the correct matching image. \\\noindent

\textbf{Proposed Method}: GeoVLM introduces a novel vision-and-language based cross-view reranking framework which learns visual reasoning to improve the top match accuracy. The approach includes pre-generating unique natural language descriptions for each image using zero-shot capabilities of VLMs. The GeoVLM uses a pre-trained vision encoder which ranks the top-10 images for each query image. Then, a trainable reranking module fuses corresponding text and image embeddings to learn to rerank the correct pair satellite image. The GeoVLM pipeline is illustrated in Figure \ref{fig:geovlm}.
\subsection{Data Generation} \label{data_genertaion}
This section details the process of generating language descriptions for cross-view images. These semantic language descriptions serve as complementary information in the proposed GeoVLM, helping to disambiguate similar-looking satellite images during the reranking process. The data generation consists of two main sub-steps: instructed Visual Question Answering (VQA) and building textual semantic descriptions.

Firstly, BLIP v2 is used as a VQA model to answer a batch of pre-defined multiple-choice questions (MCQs) for cross-view images. These MCQs are designed with the assistance of ChatGPT \cite{OpenAI2023ChatGPT} to ensure comprehensive coverage of relevant visual attributes. Using MCQs rather than open-ended questions or a single prompt like "Describe the scene" has multiple purposes. The MCQs constrain the model to focus on specific aspects of the scene that are relevant for cross-view matching while helping it avoid describing noise signals from dynamic objects, viewpoint variations, and lighting conditions. The questions are systematically designed to capture information across four main categories which are environmental context to extract information about surrounding landscape and overall setting, road features which are visible from both viewpoints such as road types, markings, and configurations and architectural elements which are mainly static over time such as buildings, colors and vegetation. 

Secondly, for each image, the questions and corresponding answers from the BLIP model are stored and then fed into a structured sentence template to build a comprehensive semantic scene description. This approach ensures descriptions maintain consistent structure, formatting, and length across all images. This controlled sentence generation process helps avoid potential hallucinations or inconsistencies that might arise from more open-ended generation methods. The complete set of MCQs and the description template are provided in Appendices A and B respectively.
\subsection{Method}
\textbf{Visual Encoder Module.} The first phase in GeoVLM pipeline is to retrieve the top-10 most similar satellite images from a database given the query image. To do so, all
reference images (denoted as $I_r$) are passed through a vision encoder using
the backbone of pre-trained Sample4Geo \cite{sample4geo} to generate
the reference embeddings, $f_{img}(I_r) = \psi_{I_r} \in \mathbb{R}^d$, where $f_{img}$ represents the image encoder function, $\psi_{I_r}$ is the output image embedding, and $d$ is the dimension of the embedding space. Similarly,
the embeddings for the given query images (denoted as $I_q$) are generated,
$f_{img}(I_q) = \psi_{I_q} \in \mathbb{R}^d$. The Sample4Geo as being the latest advancement of CVGL has shown high recall rates across various benchmarks. The backbone is a ConvNext architecture, a CNN-based model that integrates key elements of ViTs, such as the self-attention mechanism. Once the image embeddings are generated, cosine similarity is performed to rank the top-10 most similar reference images. Additionally, the corresponding natural language descriptions (captions) are retrieved for these images, which is denoted as $C_{r_i}$ for the i-th reference image and $C_q$ for the query image.  The cosine similarity and ranking are shown in equations \ref{equation1} and \ref{equation2} respectively.
\begin{equation} \label{equation1}
\text{cos}(I_q, I_{r_i}) = \frac{\psi_{I_q} \cdot \psi_{I_{r_i}}}{\|\psi_{I_q}\| \|\psi_{I_{r_i}}\|}
\end{equation}
\begin{equation} \label{equation2}
\text{top-10}(I_q) = \{ I_{r_i} \mid i \in \text{argsort}_{i=1}^{n} \left( \frac{\psi_{I_q} \cdot \psi_{I_{r_i}}}{\|\psi_{I_q}\| \|\psi_{I_{r_i}}\|} \right)[:10] \}
\end{equation}

\textbf{Reranking Module.} The second phase in the GeoVLM pipeline is to learn a joint embedding space for text and image embeddings, which is then used to produce a ranking score for the query and the \( i \)-the reference image, determining the reranking order. Given the top-10 candidates, the text embeddings for the query and reference captions are generated, denoted as \( f_{\text{text}}(C_q) \) and \( f_{\text{text}}(C_{r_i}) \), respectively, using OpenAI's text-embedding-3-small model. For each top-10 candidate, their and query's image and text embeddings are projected into a shared latent space through linear transformations. In the shared latent space, the projected image and text features are combined by element-wise summation. The combined embeddings for cross-views are then processed by a cross-embedding aligner, which consists of series of fully connected layers with layer normalization followed by ReLU activations. The output of the cross-embedding aligner is then passed through a sigmoid function to produce a final score for each query-reference canditate pair.

The final score for the query-reference image pair is calculated using the following equation:
\begin{equation} \label{equation3}
\text{S}(I_q, I_{r_i}) = \sigma\left( W \left( \mathbf{I}_q' + \mathbf{C}_q' \right) \cdot \left( \mathbf{I}_{r_i}' + \mathbf{C}_{r_i}' \right)^\top + b \right)
\end{equation}

In equation \ref{equation3}, \( \mathbf{I}_q' \) and \( \mathbf{C}_q' \) represent the projected embeddings of the query image \( I_q \) and its caption \( C_q \), respectively, while \( \mathbf{I}_{r_i}' \) and \( \mathbf{C}_{r_i}' \) are the projected embeddings for the reference image \( I_{r_i} \) and its caption \( C_{r_i} \). The term \( W \) refers to the learnable weight matrix that aligns the query and reference embeddings, and \( b \) denotes the bias term. The sigmoid function \( \sigma \) is applied to output a score ranging between 0 and 1, which represents the likelihood that the reference image corresponds to the query. The learnable parameters \( W \) and \( b \) are optimized during the training. The scoring function extends CLIP's \cite{radford2021learning} cross-modal similarity approach by projecting both modalities to a common space and combining them through element-wise addition. This approach enables text descriptions to disambiguate visually similar locations when image features alone are insufficient

\textbf{Reranking Loss.}
The reranking process is formulated as a learning-to-rank problem where the goal is to learn to distinguish the correct match from visually similar candidates. Given a query image $I_q$ and its corresponding caption $C_q$, along with a set of reference candidates $\{(I_{r_1}, C_{r_1}), ..., (I_{r_{10}}, C_{r_{10}})\}$ retrieved by the visual encoder, a similarity score $S(I_q, I_{r_i})$ is calculated for each candidate pair.

To train this reranking module, we adapt a margin-based ranking loss function \cite{nayyeri2019adaptive} that specifically addresses the challenges of cross-view matching:

\begin{equation}
\mathcal{L} = \frac{1}{|N|} \sum_{i \in N} \max(0, m - (S(I_q, I_{r_p}) - S(I_q, I_{r_i})))
\end{equation}

where $I_{r_p}$ represents the positive (correct) reference image, $m$ is the margin parameter, $N$ is the set of negative pairs in the top-10 candidates, $S(I_q, I_{r_p})$ is the similarity score for the positive pair, and $S(I_q, I_{r_i})$ is the similarity score for the $i$-th negative pair. The margin parameter $m$ is a hyperparameter defining the minimum desired difference between the positive pair score and any negative pair score. In other words, it acts as a decision boundary enforcing the separation between correct and incorrect matches in the embedding space. When $m$ is set to 1 (as in our implementation), the model is trained to ensure that the similarity score of the correct match exceeds the score of any incorrect match by at least 1 unit. This requirement creates a buffer zone that improves the reranking module's discriminative ability. Without a sufficient margin, the system might learn to make the positive score only marginally higher than negative scores, which could lead to incorrect rankings when faced with new and challenging examples.

This loss function incorporates several key design choices that make it effective for cross-view matching. Firstly, the margin parameter $m$ ensures a minimum separation between positive and negative pairs in the embedding space:
\begin{equation}
\forall i \in N: S(I_q, I_{r_p}) \geq S(I_q, I_{r_i}) + m
\end{equation}
Unlike traditional CVGL approaches which have been trained from scratch to learn new features, GeoVLM's loss function focuses to ensure the correct match ranks the highest among the similar-looking candidates. Traditional CVGL methods primarily learn generalizable feature representations through metric learning or contrastive learning objectives. However, GeoVLM benefits from the pre-trained representations and focuses exclusively on optimizing the relative ranking order of candidates that were already identified as visually similar by the base encoder. This distinction is important because:

1) The loss only activates when $S(I_q, I_{r_p}) < S(I_q, I_{r_i}) + m$, focusing solely on maintaining the correct ranking order rather than enforcing arbitrary distances between all negative pairs:
\begin{equation}
\frac{\partial \mathcal{L}}{\partial S(I_q, I_{r_i})} = 0 \text{ when } S(I_q, I_{r_p}) > S(I_q, I_{r_i}) + m
\end{equation}

2) To handle potential noise in VLM-generated descriptions,
the loss operates on the combined embedding space where
both text and image modalities contribute to the final score. The similarity function $S$ used in the loss is generated from a cross-embedding aligner that processes the combined projected embeddings.

The effectiveness of the proposed reranking approach is demonstrated through its ability to improve upon the initial top-10 retrieval by learning fine-grained discriminative features. During backpropagation, the gradient flow is concentrated on challenging cases where negative pairs score too close to the positive pair, enabling the model to identify subtle distinguishing features in both visual and textual representations. Our design ensures robust performance even in the presence of noisy or incomplete descriptions because the model learns to use complementary information from both modalities during the training process.
\section{EXPERIMENTS}
\subsection{Datasets} \label{dataset}
The performance of GeoVLM is evaluated on three main benchmark datasets: The CVUK, VIGOR and University-1652 datasets. \\\noindent
\textbf{VIGOR}: The VIGOR dataset contains 105,214 street-view panorama images and 133,482 satellite images from four different states in the U.S; New York, San Francisco, Seattle and Chicago. The VIGOR contains ground level images that may belong to random spots within the target area and the images may not be aligned with the center of satellite images. The VIGOR dataset offers two evaluation configurations: \textit{same-area} and \textit{cross-area}. The same-area setting of VIGOR dataset means that the training and testing of the chosen model are performed in the same geographical regions where the cross-area means that the training is performed in New York and Seattle and testing is performed in San Francisco and Chicago. \\\noindent
\textbf{CVUK}: The CVUK dataset consists of driving footage captured from an ego-vehicle traveling through Liverpool, London, and Woking. The data was collected in July 2023, December 2023, and January 2024 and covers different lighting, seasonal, and environmental conditions. It includes approximately 8 hours of driving footage with GPS coordinates and corresponding aerial views extracted from Google Maps. The aerial images are not necessarily center aligned but they are spatially aligned with the given GPS positions. The CVUK data is sampled and the satellite images are at least 100 meters apart from each other. The CVUK dataset includes $\approx$3k satellite and query image pairs. The satellite and query images are $640 \times 640$ and $1920 \times 1080$ pixels, respectively. The CVUK offers a real-life representation of wide range of urban, suburban and rural driving conditions which include varying traffic densities, diverse architectural landscapes, and different weather conditions of the UK's climate. The time and season of each trip is presented in Table \ref{tab:time_season}.
\\\noindent
\begin{table}[h]
\centering
\caption{Time and Season of Trips in the CVUK Dataset}
\label{tab:time_season}
\footnotesize
\begin{tabular}{cc}
\hline
\textbf{Trip} & \textbf{Time and Season} \\
\hline
Woking to London & January 2024, 13:00 to 14:30 \\
Liverpool to Woking 1 & December 2023, 14:00 to 17:00 \\
Liverpool 1 & December 2023, 15:00 to 16:00 \\
Liverpool 2 & December 2023, 16:00 to 18:30 \\
Liverpool to Woking 2 & July 2023, 19:00 to 22:00 \\
Woking to Liverpool & July 2023, 20:00 to 23:00 \\
\hline
\end{tabular}
\end{table}

\textbf{University-1652}: Unlike the CVUK and VIGOR, University-1652 introduced by \cite{zheng2020university} is a multi-view and multi-source dataset focused on drone-based geo-localisation, containing satellite-view, drone-view and ground-view images. The dataset consists of 1,652 buildings from 72 universities worldwide.  The training set includes 50,218 images of 701 buildings from 33 universities, while the testing set contains buildings from 39 different universities. University-1652 is a relevant benchmark to assess the benefits of the proposed reranking based GeoVLM approach as it provides multiple views of the same location, which can help validate the effect of language descriptions across different viewpoints.
\subsection{Evaluation Metric}
To evaluate the proposed approach, recall accuracy at top K (R@K) is used for K=1, 5 and 10 on the VIGOR dataset. However, the main evaluation metric for GeoVLM is the improvement in R@1 and R@5 as the reranking is applied from R@10. For the University dataset, R@1 and Average Precision (AP) are used as the primary evaluation metrics. For the CVUK benchmark, due to its challenging nature with sparse features in highway driving scenarios, the evaluation adopts a position-based methodology. Specifically, the evaluation determines if the retrieved images in the R@K fall within predefined positional error thresholds (0.0km and 0.5km) of the correct location. The positional accuracy is measured using the Haversine distance, which calculates the shortest distance between two points on the Earth's surface based on their latitude and longitude coordinates. Results are reported in kilometers (km) to quantify the geographical precision of the retrieval system under these challenging conditions.
\subsection{VIGOR Evaluation}
GeoVLM is evaluated for both the same-area and cross-area settings of the VIGOR. For each setting the corresponding pre-trained vision encoder is used to identify the initial ranks. The reranking module is trained with the queries and their corresponding top-10 retrievals where the exact correct pair is found in top-10. The training of reranking module naturally includes the semi-positive and false-positive images which is reported as the main challenge in prior works \cite{sample4geo}, \cite{Zhu2022TransGeo}. The semi-positive images are those which are visually very similar to the ground truth and includes overlapping regions from the ground truth but geographically taken from different positions that they are not center aligned with the query image. The GeoVLM uses the textual descriptions along with the image features to learn to disambiguate the ground truth from semi or false positive images. The query and satellite images are resized to $384 \times 768$ and $384 \times 384$ pixels, respectively. The image and text embeddings are 1024 and 1536 dimensional respectively. The results illustrated in Table \ref{tab:comparison_same_cross} show that GeoVLM outperforms the SAFA \cite{safa}, TransGeo \cite{Zhu2022TransGeo}, SAIG-D \cite{saig} by improving the performance of  Sample4Geo \cite{sample4geo} for R@1 by 9.9\% and 6.9\% for same-area and cross-area respectively. The results also demonstrate that the use of language descriptions allows GeoVLM to learn the association between the textual descriptions and the image features enabling for a better geometric and visual understanding of the scene. Considering the cross-area setting, the improvement in R@1 is less compared to the same-area setting because the R@10 is less. This also suggests that GeoVLM's improvement is dependent on the pretrained backbone's accuracy in identifying the correct image within top-10.    
\subsection{CVUK Evaluation}
In order to further evaluate the generalisability and applicability to real-world scenarios of GeoVLM and compare it with other SOTA approaches like TransGeo \cite{Zhu2022TransGeo} and Sample4Geo \cite{sample4geo} experiments are performed using the CVUK dataset which include query images from various environment, varying light conditions and across different seasons. As CVUK is a challenging dataset positional error threshold is introduced to validate the retrieval accuracy for different positional errors. Additionally, due to the relatively small number of samples the CVUK is used for fine-tuning and validating the pre-trained models. The first 3 rows of the Table \ref{tab:cvuk} presents the results for experiments where the VIGOR pre-trained models (SampleGeo and TransGeoare tested on the CVUK data for different positional errors. The best accuracy is achieved by the Sample4Geo \cite{sample4geo} backbone and the GeoVLM improved the results by an average of 16\% and 7\% for different positional errors for R@1 and R@5 respectively. Moreover, the pre-trained VIGOR models are fine-tuned with CVUK data and re-validated on VIGOR cross-area setting. The experiments show a significant drop in the performance of SOTA methods when they are fine-tuned with real-world scenarios. However, the improvement of GeoVLM is still reported as 14\% and  5\% for R@1 and R@5 respectively.

\begin{table}[!t]
\caption{Performance Comparison of GeoVLM and Other SOTA Approaches in the VIGOR Benchmark}
\centering
\setlength{\tabcolsep}{3.5pt}
\renewcommand{\arraystretch}{1.3}
\footnotesize
\begin{tabular}{l ccc ccc}
\hline
\textbf{Approach} & \multicolumn{3}{c}{\textbf{SAME}} & \multicolumn{3}{c}{\textbf{CROSS}} \\
\cline{2-4} \cline{5-7}
 & \textbf{R@1} & \textbf{R@5} & \textbf{R@10} & \textbf{R@1} & \textbf{R@5} & \textbf{R@10} \\
\hline
SAFA & 33.93 & 58.42 & 68.12 & 8.20 & 19.59 & 26.36 \\
TransGeo & 61.48 & 87.54 & 91.88 & 18.99 & 38.24 & 46.91 \\
SAIG-D & 65.23 & 88.08 & -- & 33.05 & 55.94 & -- \\
Sample4Geo & 77.86 & 95.18 & 97.21 & 61.70 & \textbf{83.50} & 88.00 \\
GeoVLM & \textbf{85.64} & \textbf{96.18} & \textbf{97.21} & \textbf{71.98} & 81.87 & \textbf{88.00} \\
\hline
\end{tabular}
\label{tab:comparison_same_cross}
\end{table}

\begin{table}[!t]
\caption{Performance Comparison of Different Approaches Across Tasks and Error Thresholds}
\centering
\footnotesize
\begin{tabular}{lccc}
\hline
\textbf{Approach} & \textbf{R@1} & \textbf{R@5} & \textbf{R@10} \\
\hline
\multicolumn{4}{l}{\textbf{VIGOR $\rightarrow$ CVUK (Error Threshold: 0.0 km)}} \\
TransGeo \cite{Zhu2022TransGeo} & 3.04 & 7.21 & 10.38 \\
Sample4Geo \cite{sample4geo} & 7.24 & 9.33 & 12.50 \\
\textbf{GeoVLM} & \textbf{8.78} & \textbf{10.16} & \textbf{12.50} \\
\hline
\multicolumn{4}{l}{\textbf{VIGOR $\rightarrow$ CVUK (Error Threshold: 0.5 km)}} \\
TransGeo \cite{Zhu2022TransGeo} & 5.21 & 16.13 & 24.56 \\
Sample4Geo \cite{sample4geo} & 11.23 & 24.56 & 31.05 \\
GeoVLM & \textbf{13.50} & \textbf{26.90} & \textbf{31.05} \\
\hline
\multicolumn{4}{l}{\textbf{CVUK $\rightarrow$ VIGOR (No Error Threshold)}} \\
TransGeo \cite{Zhu2022TransGeo} & 14.13 & 28.36 & 32.14 \\
Sample4Geo \cite{sample4geo} & 36.07 & 45.66 & 73.90 \\
GeoVLM & \textbf{41.21} & \textbf{47.90} & \textbf{73.90} \\
\hline
\end{tabular}
\label{tab:cvuk}
\end{table}

\subsection{University-1652 Evaluation}
GeoVLM is also validated on the University-1652 benchmark using the Drone2Sat and Sat2Drone settings. The University dataset is a unique evaluation scenario as it includes multiple correct images for a given query image. To ensure a more rigorous assessment of our reranking module a conservative sampling approach is implemented during both training and testing. Specifically, the batch is sampled such that only one positive reference image is included in the reference dataset for each query image. The reason for this sampling strategy is due to the already high recall rates achieved by existing SOTA methods DWDR \cite{wang2022learning}, MBF \cite{deuser2023orientation}, and Sample4Geo \cite{sample4geo}) on this benchmark, allowing us to more clearly isolate and evaluate the specific contribution of the proposed reranking module.
Additionally, to ensure comprehensive evaluation, the same query image is processed multiple times, each time with a different correct pair from its set of matching images. This approach ensures that all possible correct pairs are being evaluated, providing a more complete assessment of GeoVLM's ability to recognize the same location across different viewpoints and conditions. 
The results illustrated in Table \ref{tab:performance_comparison} demonstrate that GeoVLM outperforms the mentioned SOTA approaches across all evaluation metrics. In the Drone2Sat setting, GeoVLM achieves 93.15\% for R@1 and 95.23\% for AP, representing improvements of 0.5\% and 1.42\% respectively over Sample4Geo \cite{sample4geo}, the previous best-performing method. In the Sat2Drone setting, GeoVLM achieves 97.18\% for R@1 and 93.36\% for AP, surpassing Sample4Geo by 2.04\% and 1.97\% respectively.

\begin{table}
\caption{Performance comparison of different approaches on Drone2Sat and Sat2Drone metrics.}
    \centering
    \begin{tabular}{lcccc}

        \hline
        \multirow{2}{*}{Approach} & \multicolumn{2}{c}{Drone2Sat} & \multicolumn{2}{c}{Sat2Drone} \\
        \cline{2-5}
        & R@1 & AP & R@1 & AP \\
        \hline
        DWDR \cite{wang2022learning} & 86.41 & 88.41 & 91.30 & 86.02 \\
        MBF \cite{deuser2023orientation} & 89.05 & 90.61 & 92.15 & 84.45 \\
        Sample4Geo  \cite{sample4geo} & 92.65 & 93.81 & 95.14 & 91.39 \\
        GeoVLM & \textbf{93.15} & \textbf{95.23} & \textbf{97.18} & \textbf{93.36} \\
        \hline
    \end{tabular}
    \label{tab:performance_comparison}
\end{table}

\subsection{Visualisation}
\begin{figure}[h]
    \centering
    \includegraphics[scale=0.6]{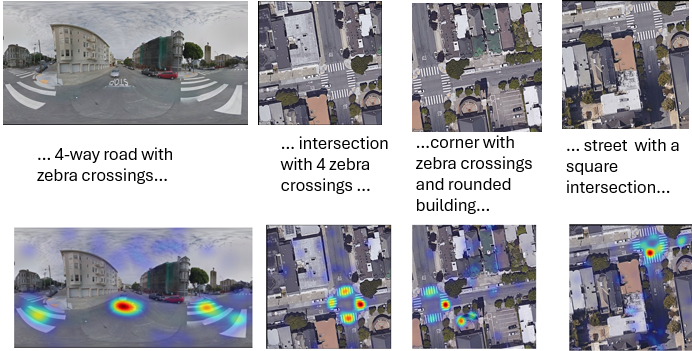}
    \caption{Visualization of cross-view images with their corresponding captions and the attention of their text tokens on the relevant part of the image.}
    \label{blip_vqa}
\end{figure}

\begin{figure}[h]
    \centering
    \includegraphics[scale=0.4]{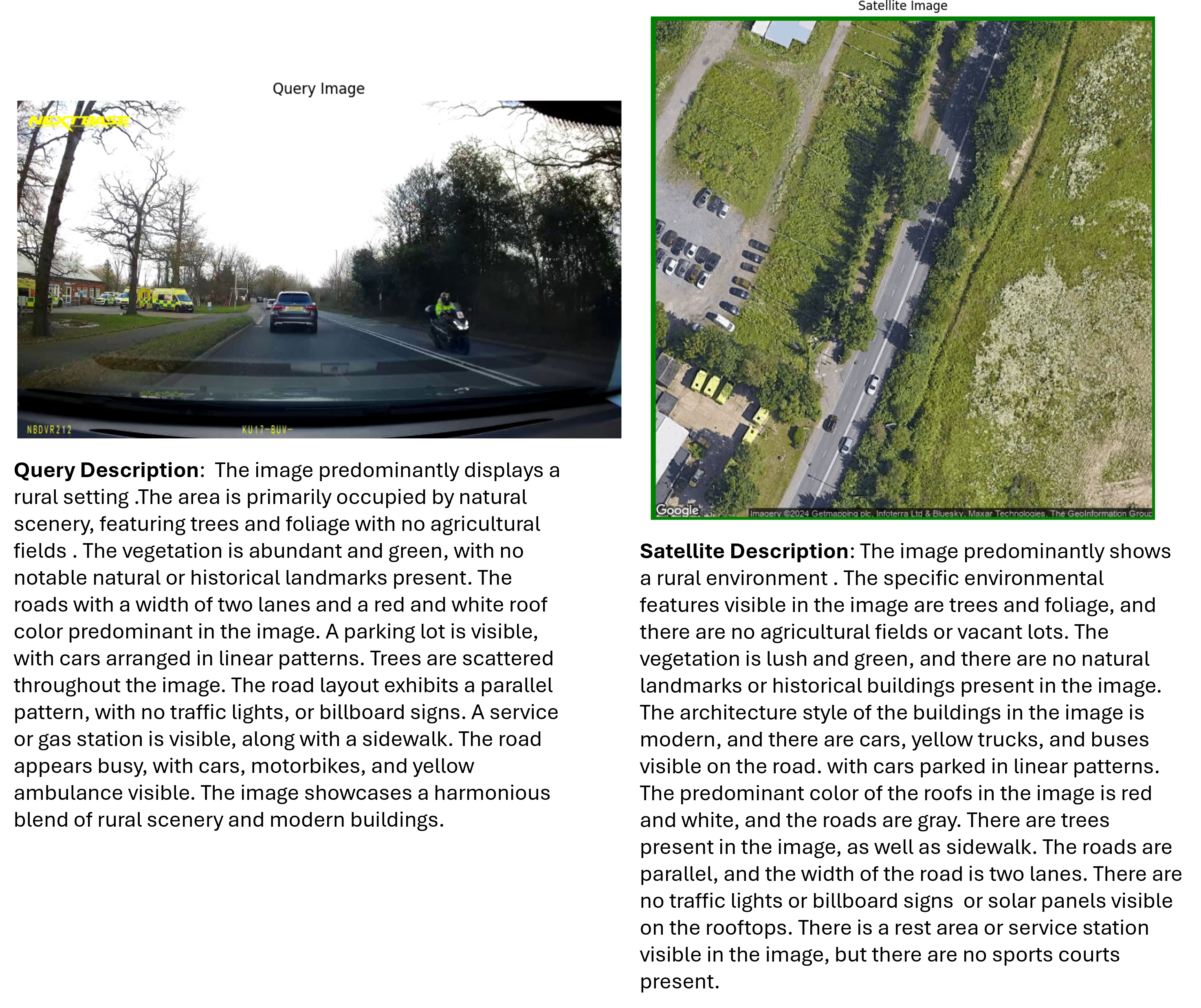}
    \caption{Visualization of a correct match from the CVUK dataset with corresponding scene descriptions. The descriptions were generated using our MCQ-based VQA approach, where BLIP responded to 30 structured questions about the scene.}
    \label{correct_match}
\end{figure}

\begin{figure}[h]
    \centering
    \includegraphics[scale=0.35]{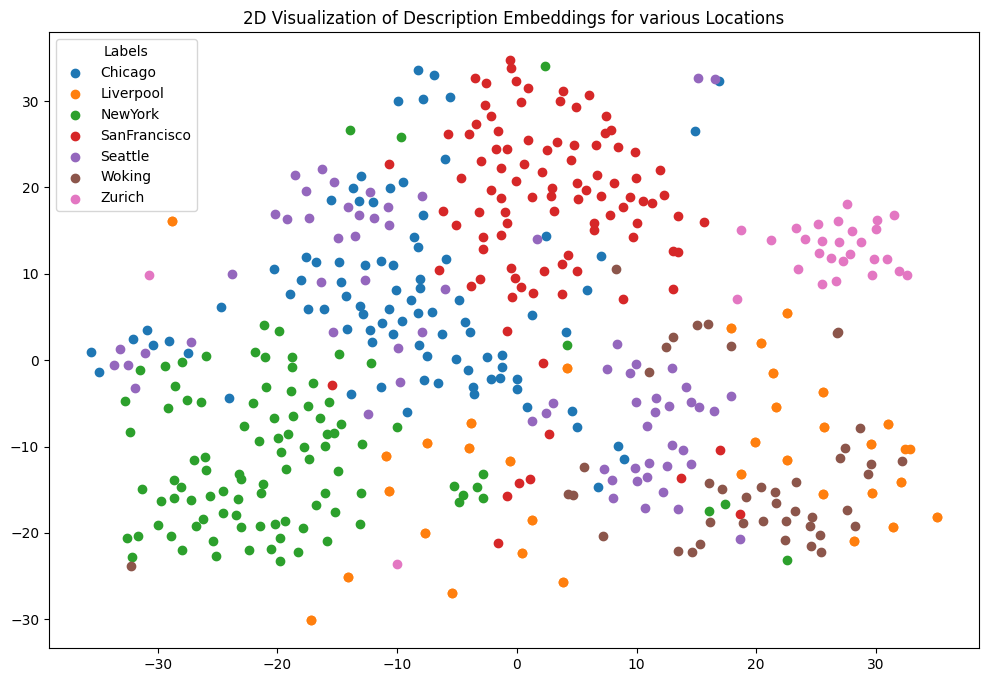}
    \caption{Visual clusters of image descriptions for various
locations in the CVUK and VIGOR dataset. Each color represents a unique city or state, highlighting the natural
clustering of the scene description embeddings in a reduced-
dimensional space}
    \label{fig:visual_clusters}
\end{figure}

To illustrate the use of natural language descriptions in reranking cross-view images and show the importance of human level reasoning, Figure \ref{blip_vqa} is presented. The Figure \ref{blip_vqa} illustrates a query image and its corresponding top-3 retrievals with a snapshot of their captions which enabled the model to rerank the ground truth as top-1. The heat-maps of each are also displayed indicating the focused regions for corresponding text tokens generated by the BLIP model. These visualizations demonstrate the VLM's targeted attention for extracting relevant scene attributes for the cross-view images. On the other hand, Figure \ref{correct_match} demonstrates a correctly identified match from the CVUK test dataset, along with the scene descriptions for both the query and top-1 match images. The scene descriptions are generated by the MCQ-based VQA approach detailed in Section \ref{data_genertaion}, where each image is processed with a set of 30 carefully designed multiple-choice questions. For each image, BLIP2 is prompted with questions covering environmental context, road features, architectural elements, and other distinguishing characteristics. The model's responses to these MCQs are then inserted into a structured template that combines them into a coherent paragraph, creating a comprehensive scene description. The figure illustrates how semantically similar captions can improve the reranking performance through the zero-shot capabilities of VLMs. The text modality helps explain invariant scene attributes that persist across different viewpoints. The proposed GeoVLM effectively distinguishes between critical spatial features (such as road layouts and building arrangements) over dynamic objects benefiting from its reasoning capabilities about scene structure rather than relying on superficial visual similarities. 
In addition to that, Figure \ref{fig:visual_clusters} plots a sample batch of image description embeddings in a reduced dimensional space from each city and state covered in the CVUK and VIGOR dataset which demonstrates the discriminative power of the zero-shot GeoVLM approach across various geographic locations. \\\noindent

\subsection{Stability Check for Caption Generation}
To evaluate the consistency and reliability of the VLMs, a stability check was conducted with 20,000 randomly selected images from the CVUK and VIGOR, using the same MCQ set and repeating the VQA process twice. Table \ref{tab:stability_check_results} presents the key metrics from this stability analysis.

\begin{table}[h]
    \centering
    \renewcommand{\arraystretch}{1.0}
    \caption{Stability Check Results for LLM Descriptions}
    \label{tab:stability_check_results}
    \small
    \begin{tabular}{lc}
        \hline
        \textbf{Parameter} & \textbf{Value} \\
        \hline
        Cosine Similarity & 0.83 \\
        Jaccard Similarity & 0.44 \\
        Mean Length & 185.49 \\
        Standard Deviation Length & 14.76 \\
        VQA Response Time & 8.3s \\
        \hline
    \end{tabular}
\end{table}

As reported in Table \ref{tab:stability_check_results}, the cosine similarity of the descriptions was 0.83, indicating high semantic similarity. The Jaccard similarity, which measures the number of words overlapping between two sentences over the union of sentences, is at 0.44, showing moderate overlap in vocabulary and suggesting some variability in the words used. The mean description length was 185.49 words with a standard deviation of 14.76, demonstrating consistent but variable length. The average VQA response time was 8.3 seconds, highlighting potential limitations of LLMs and VLMs for time-critical tasks in automated driving such as localisation.
\section{DISCUSSION \& CONCLUSION}
In this paper, we presented GeoVLM, an explainable vision-language reranking approach for the CVGL task which improves the top-match accuracy by integrating human level visual reasoning into the cross-view image retrieval task. We compared GeoVLM with the SOTA methods through the VIGOR, CVUK and the University-1652 benchmarks. The CVUK dataset is a collection of journeys across three major cities of the UK at various seasons and times of the day, providing a challenging and realistic evaluation environment for geo-localisation approaches.

The paper highlighted a critical gap in the current approaches, which fail to identify the correct image pair among similar looking scenes as well as reaching low recall rates in realistic and challenging driving scenarios such as on a highway or a rural areas with minimal scenery changes and sparse features. The proposed GeoVLM improved the retrieval performance in these situations by using its large vocabulary to identify and interpret subtle cues that remain consistent across different viewpoints. Our results demonstrate how natural language descriptions help to identify similar looking scenes through visual reasoning capabilities. The performance improvements demonstrated by GeoVLM across all three benchmark datasets VIGOR, CVUK, and University-1652 validated that language descriptions can provide a way of disambiguation for CVGL. For the VIGOR dataset, GeoVLM improved the recall rates in both same-area and cross-area settings. This showed that the language-based reranking approach captures semantic relationships between cross-view images that purely visual features might miss. Similarly, for the University-1652 dataset, GeoVLM demonstrated consistent improvements across both Drone2Sat and Sat2Drone configurations.

The most challenging evaluation scenario, the CVUK dataset, provided valuable insights into real-world applicability. The evaluation results highlighted a significant limitation of current vision-based approaches, which achieve notably low recall rates in challenging driving environments such as highways with sparse distinguishing features or rural areas with repetitive scenery. Traditional vision-only methods struggle to discriminate between visually similar locations in these scenarios, often resulting in incorrect matches or high positional uncertainty. Although, additional text modality with reranking approach demonstrated improvements across all datasets, it has limitations that currently restrict its integration into real-time localisation systems for automated vehicles. A significant constraint is that the scene descriptions are pre-generated offline, and no VLM is used during training or testing. This implementation choice was necessary to mitigate the high latency associated with VLM inference. The visual question answering and semantic description generation processes required significantly longer response times compared to existing vision-only approaches, as evidenced by our stability check results. Additionally, the computational cost for running a VLM server is substantially higher than vision-based approaches, raising concerns about practical deployment in resource-constrained environments. Considering all these factors of multi-modal large language models, their integration to real-time localisation needs further study. In future work, we plan to implement the language-based reranking approach on a physical robotic vehicle to further investigate the suitability of the approach for real-time integration in automated navigation systems.  
\section*{ACKNOWLEDGMENT}
The first author thanks research funding support from UK Engineering and Physical Science Research Council. The authors acknowledge the use of AI-based tools in preparting the manuscript. Specifically, BLIP VQA model was used to generate the descriptions illustrated in Figure \ref{blip_vqa} and \ref{correct_match}.

%
\clearpage
\twocolumn
\section*{A. Appendix}

\subsection*{A.1. Multiple-Choice Questions for Cross-View Language Generation}

We used a set of 30 carefully designed multiple-choice questions to generate comprehensive and consistent cross-view language descriptions. These questions systematically capture key visual elements from both ground and aerial imagery.

\begin{table}[h]
\scriptsize
\centering
\caption{Multiple-Choice Questions (Part 1)}
\begin{tabular}{|c|p{5.8cm}|}
\hline
\textbf{No.} & \textbf{Question (Options)} \\
\hline
Q1 & What is the predominant environment? (urban/suburban/rural/highway/industrial/natural/dense forestation/water body/mixed) \\
\hline
Q2 & What type of road layout is visible? (grid pattern/winding roads/roundabout/dead-end streets/highway/none/mixed patterns) \\
\hline
Q3 & What specific environmental features are visible? (Residential buildings/Commercial areas/Factories/Farms/Green spaces/Parks/Rivers/Lakes/Forests/Beaches/Cliffs/Hills/Open fields) \\
\hline
Q4 & What kind of distinct road features are present? (none/simple intersections/complex junctions/overpasses/roundabouts/traffic circles) \\
\hline
Q5 & What types of buildings are most common? (residential houses/apartment buildings/commercial buildings/industrial facilities/public buildings/mixed/no buildings) \\
\hline
Q6 & What is the condition of the vegetation? (None/dense forests/parklands/sparse vegetation/agricultural fields/barren land/ornamental gardens) \\
\hline
Q7 & What distinctive features are present? (None/Natural Landmarks/Historical Buildings/Modern Structures/Sporting Facilities/Water Bodies/Parks/Urban Art/Monuments) \\
\hline
Q8 & What is the architecture style of the buildings? (None/traditional/modern/industrial/mixed/historical) \\
\hline
Q9 & What transportation features can be seen? (None/train tracks/airports/ports/tram lines/bus stations) \\
\hline
Q10 & What kind of large, open spaces are there? (None/fields/empty lots/forests/car parks/urban squares/golf course/public garden/playgrounds/sports field) \\
\hline
Q11 & What is the overall layout of the area? (organized/disorganized/mixed/regular pattern/irregular pattern/none/chaotic) \\
\hline
Q12 & What unique patterns appear in roads or buildings? (none/linear patterns/radial patterns/grid patterns/irregular patterns/circular patterns) \\
\hline
Q13 & What is the predominant color of the roofs? (red/brown/grey/white/green/other/none/multi-colored) \\
\hline
Q14 & What is the predominant color of the roads? (black/grey/red/yellow/other/none/multi-colored) \\
\hline
Q15 & What other notable color features are present? (green areas/water bodies/colored buildings/sports fields/none/colorful gardens) \\
\hline
\end{tabular}
\end{table}

\begin{table}[h]
\scriptsize
\centering
\caption{Multiple-Choice Questions (Part 2)}
\begin{tabular}{|c|p{5.8cm}|}
\hline
\textbf{No.} & \textbf{Question (Options)} \\
\hline
Q16 & What type of main road is visible? (none/single-lane road/multi-lane road/highway/expressway) \\
\hline
Q17 & What road markings are present? (None/Zebra crossings/Chevrons/White lines/Yellow lines/Double yellow lines/Arrows/Stop lines/Crosswalks/Bicycle lanes) \\
\hline
Q18 & What are the predominant colors of the road markings? (White/Yellow/Red/Blue/Green/None/Other/Multi-colored) \\
\hline
Q19 & Are any road structures visible? (None/Bridge/Underpass/Overpass/Tunnel/Flyover/Pedestrian crossing bridge/Roundabout/Highway interchange) \\
\hline
Q20 & How would you describe the orientation of the roads? (Straight highway/Single road/Multiple parallel roads/Roads converging/Roads diverging/Intersection/Roundabout) \\
\hline
Q21 & What are the predominant types of surrounding vehicles? (Cars/Trucks/Bicycles/Motorcycles/Public Transport/None) \\
\hline
Q22 & What is the directional layout of the road junction? (none/left turn only/right turn only/both turns/four-way intersection/roundabout/multiple directions) \\
\hline
Q23 & What is the width of the road? (None/narrow/medium/wide/multiple lanes/variable widths) \\
\hline
Q24 & Are there any traffic lights present along the road? (yes/no) \\
\hline
Q25 & Are there any billboard signs indicating directions or destinations? (Yes/No) \\
\hline
Q26 & Is there a rest area or service station visible? (yes/no) \\
\hline
Q27 & What type of service facility is visible? (None/Petrol station/Supermarket/Restaurant/Hotel) \\
\hline
Q28 & Are any sports courts visible? (None/basketball/tennis/football) \\
\hline
Q29 & Does the road have a hard shoulder or emergency lane? (yes/no) \\
\hline
Q30 & Is there a pedestrian area like a sidewalk or footpath? (yes/no) \\
\hline
\end{tabular}
\end{table}

\subsection*{A.2. Description Generation Template}

After collecting responses to all questions, we generated structured descriptions using the following template:

The image shows a [Q1] area with a [Q2] road layout, featuring [Q3] such as [Q4]. The buildings are predominantly [Q5], with vegetation described as [Q6]. Distinctive features include [Q7], and the architecture style is [Q8]. 

Transportation features include [Q9], with open spaces like [Q10]. The area is [Q11] in layout, with [Q12] patterns. The roofs are predominantly [Q13] in color, while the roads are [Q14] with [Q15]. 

The main road visible is a [Q16], with road markings such as [Q17] in [Q18]. Road structures include [Q19], and the road orientation is [Q20]. Surrounding vehicles are mainly [Q21]. The junction allows [Q22] traffic flow, with a road width of [Q23]. 

Traffic lights: [Q24], billboard signs: [Q25]. A service station is [Q26] visible, offering [Q27]. Sports courts ([Q28]) are visible, hard shoulder: [Q29], and pedestrian area: [Q30].

\clearpage
\twocolumn
\newpage

\section{Biography Section}
\vspace{11pt}

\begin{IEEEbiography}[{\includegraphics[width=1in,height=1.25in,clip,keepaspectratio]{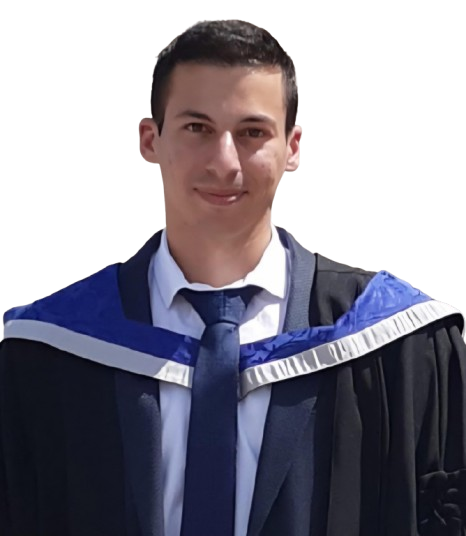}}]{Barkin Dagda}
received the B.Eng. degree in Mechanical Engineering and the M.Sc. degree in Electronic Engineering from the University of Surrey, Guildford, U.K., in 2021 and 2022, respectively. He is currently pursuing the Ph.D. degree in Automotive Engineering at the University of Surrey, within the Connected and Autonomous Vehicles (CAV) Lab. His PhD  focuses on vision-based foundation models for autonomous vehicles. With experience in academia and industry, he has worked on AI-driven robotics, navigation, and automation
\end{IEEEbiography}

\begin{IEEEbiography}[{\includegraphics[width=1in,height=1.25in,clip,keepaspectratio]{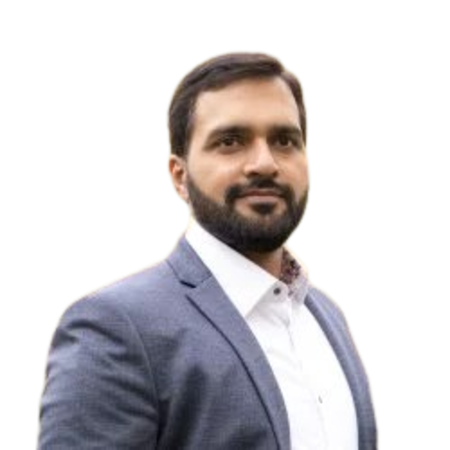}}]{Muhammad Awais}
(Member, IEEE) received the B.Sc. degree in mathematics and physics from AJK University, Muzaffarabad, Pakistan, in 2001, the second B.Sc. degree in computer engineering from University of Engineering and Technology, Taxila, Pakistan, in 2005, the M.Sc. degree in signal processing and machine intelligence, and the Ph.D. degree in machine learning from the University of Surrey, Guildford, U.K., in 2008 and 2011, respectively. He is currently a Senior Lecturer of trustworthy and responsible AI with the Surrey Institute for People-Centred Artificial Intelligence and Centre for Vision, Speech and Signal Processing, University of Surrey. His research interests include machine learning, deep learning, representation learning, multi-modal representation learning, NLP, audio-visual analysis, medical image analysis, and computer vision.
\end{IEEEbiography}

\begin{IEEEbiography}[{\includegraphics[width=1in,height=1.25in,clip,keepaspectratio]{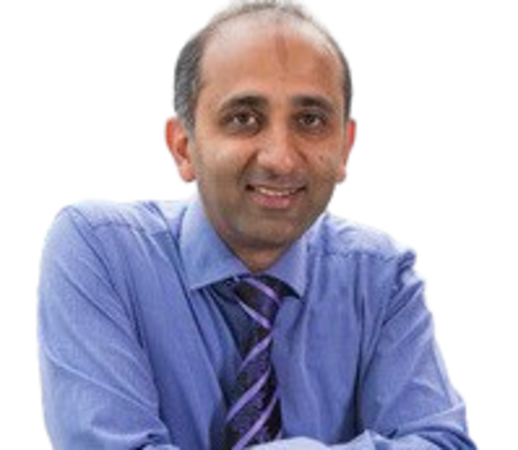}}]{Saber Fallah}
is a Professor of Safe AI and Autonomy, specializing in autonomous vehicles, self-driving cars, and autonomous robots. His research focuses on the safe deployment of AI technologies, with particular emphasis on intelligent transportation systems and autonomous decision-making. As the Director of the Connected Autonomous Vehicles Lab (CAV-Lab) at the University of Surrey, he leads a team dedicated to integrating cognitive adaptability into AI systems. This research enhances the safety and decision-making capabilities of autonomous vehicles, enabling them to better understand their environment and human behaviour. Prof. Fallah has led a range of research initiatives funded by the UK and European governments, as well as industry (EPSRC, Innovate UK, H2020, KTP).
\end{IEEEbiography}

\vfill

\end{document}